\documentclass[twocolumn,showpacs,showkeywords,preprintnumbers,amsmath,amssymb]{revtex4}

\usepackage{graphicx}
\usepackage{dcolumn}
\usepackage{bm}
\usepackage{epsfig} 
\usepackage{csquotes}

\begin{document}

\preprint{}

\title{Combinational neural network using Gabor filters for the classification of handwritten digits  }

\author{N. Joshi}
\email{njoshi@fias.uni-frankfurt.de}

\affiliation{%
Frankfurt Institute for Advanced Studies (FIAS),\\
600438 Frankfurt, Germany 
}
\affiliation[Also at]{Goethe University, Frankfurt am Main.}

\date{September 10, 2017}

\begin{abstract}

A classification algorithm that combines the components of k-nearest  neighbours and multilayer neural networks has been designed and tested.
With this method the computational time required for training the dataset has been reduced substancially.
Gabor filters were used for the feature extraction to ensure a better performance. 
This algorithm is tested with MNIST dataset and it will be integrated as a module in the object recognition software which is currently under development.

\end{abstract}

\keywords{Gabor filter, Neural Network, Machine learning, Nearest Neighbour, Computer vision}
\pacs{07.05.Mh,42.30.Tz, 42.30.Sy }

\maketitle

\section{\label{sec:level1}Introduction\protect\\
 }

Deep learning algorighms are gaining fame with its performance on the large datasets.
They rely on the availability of  huge data, thereby minimizing the influence of few irregular samples from dataset.
A classification algorithms that combines the k-nearest nearest neighbours and multi layer perceptron (MLP) network has been designed and tested.
A fusion of parametric and non-parametric approach produces better learning results with a few trainning examples.
With this combinational approach, we have also managed to reduced a substancial amount dataset required to train the network.
This algorithm has been tested on the well known dataset of handwritten digits from  MNIST \cite{MNIST}.
This approach is known as \emph{meta learning} \cite{MetaLearn}.

\section{\label{sec:level1}Gabor filters and feature extraction\protect\\
 }

\begin{figure}[hhh]
    \centering
    \includegraphics[width=8cm]{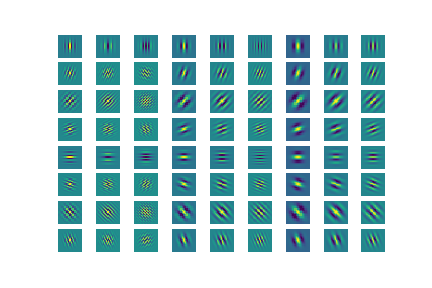}
    \caption{Gabor filter bank created by varying $\sigma$, frequency ($\nu = 2 \pi / \lambda$) and rotation ($\theta$). The intensities are false colour coded for better visual purpose.
      }
    \label{fig_1}
\end{figure}

Gabor filters are the natural choice to extract features as they produce a respose similar to the human visual system \cite{Gabor_1}. 
A two-dimentional Gabor function is written as

\begin{equation}
g \big( x,y; \lambda,  \theta, \psi, \theta, \gamma \big) = exp \Big(  - \frac{x'^ 2 + \gamma^2 y'^ 2 }{2 \sigma^2}\Big)    exp \Big( i  \big( 2 \pi \frac{x'}{ \lambda} + \psi \big)    \Big) ,
\end{equation}

where 
\begin{eqnarray*}
x'&=& x~cos(\theta)+y~sin(\theta),\\
y'&=& - x~sin(\theta)+y~cos(\theta),
\end{eqnarray*}

and $ \lambda$ is the wavelength, $ \theta$ is  the orientation, $\psi$  is the phase offset, $\sigma$ is the standard deviation of the Gaussian envelope and $\gamma$ is the spatial aspect ratio of the Gabor function.
A Gabor filter bank is created by the choice of different parameters.
In this case three values of  $\sigma$ and $ \lambda$ were chosen with eight rotations ($\theta$ 's) equally spaced between $[0,\pi]$ (Total $3 \times 3 \times 8 = 72$ ).
Fig.\ref{fig_1} shows pictures of different gabor filters in 2D with false colour coded intensities.

\begin{figure}[hhh]
    \centering
    \includegraphics[width=8cm]{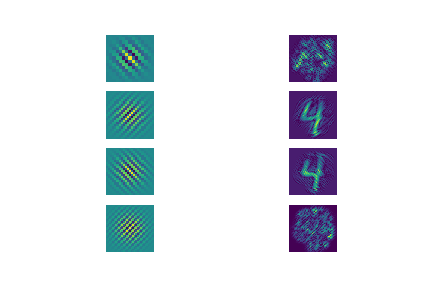}
    \caption{An example of four Gabor filters applied to the sample image.
      }
    \label{fig_2}
\end{figure}

Each of the gabor filter is convolved with the sample image (see example Fig.\ref{fig_2}) to get the respose matrix.
All of these response matrices then converted to form a feature vector.

\section{\label{sec:level1}Combined network Model\protect\\
 }

Gabor filter can be used for boundary detection or texture segmentation.
In the case of  handwritten digits recognition the inner structure is ignored. 
Only frequencies ($\nu = 2 \pi / \lambda $) and the values of Gaussian spread $\sigma$ responsible for detecting the form of figure were used \cite{Gabor_2,Gabor_3}.
One should adjust these values according to image properties like the size and depth.
Please also note that, each image manipulation library uses different methods to read the raw data from the image file, giving rise to conflicting results.
In this case, an image was imported as a float array and normalized to avoid any floating point errors.
A feature vector can be formed by calculating local energy, mean amplitude, entropy or phase amplitude.
We used the information entropy and the energy to form a vector of $[1 \times 144]$ dimension.
The Shannon information entropy is defined as

\begin{equation}
H = - \sum_{k}^{} p_k log_2(p_k) ,
\end{equation}

where $k$ is the number of gray levels and $p_k$ is the probability associated with gray level $k$ \cite{Shannon}.

 \begin{figure}[hhh]
    \centering
    \includegraphics[width=8.cm]{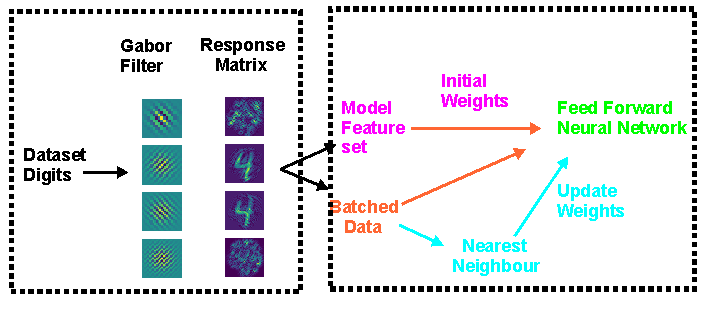}
    \caption{Schematics of neural network model.
      }
    \label{fig_3}
\end{figure}

Fig.\ref{fig_3} shows the scheme of the algorithm.
Models of digits were created by taking an average of a certain number of randomly chosen images from the MNIST dataset.
The feature vector extracted from this model was used to initialize an MLP network.
Then a batch of dataset from trainning set was fed.
When the cost function exceeds the pre-determenined value calculated from covariance matrix, a newly calulated \emph{centroid}  was used to update the weights of the neural network. 
Thus, with this strategy, a faster recognition of particular image was achieved using lesser number of trainning samples.

\section{\label{sec:level1}Experiments and Results\protect\\
 }

It was observed that the efficiency of the algorithm depends on the number of elements including the Gabor parameters as well as standard deviation allowed to update weights of the network.
In the first stage, the Gabor parameters were varied in such a way that the distances between vectors generated from model digits were optimized.

 \begin{figure}[hhh]
    \centering
    \includegraphics[width=8.cm]{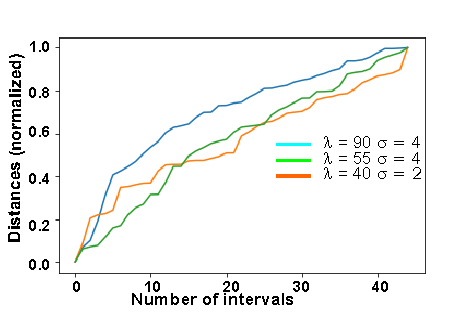}
    \caption{Minkowski distances distributed over invervals for different values of frequency  and $\sigma$.
      }
    \label{fig_4}
\end{figure}

Fig.\ref{fig_4} shows the normalized distribution of Minkowski distances between each digits. 
The Minkowski distance between two points $x_1$ and $x_2$ of order $p$ is defined as 

\begin{equation}
 d = \big( \sum_{i=1}^{n} \left( \left| x_1 - x_2 \right| \right) ^{p} \big) ^{1/p}.
\end{equation}

For $p=2$, it reduces to the Euclidean distance.

It is desired that for best classification the centroids should be as far as possible with equal distances between them.
Inspite of that, one observes that the distance between digits 1 and 7 is much less than the distance between 1 and 8 due to their similarity.
When frequencies and $\sigma$ are optimized we get the distribution depicted by blue solid line (see Fig.\ref{fig_4}).
The number of interval denotes the vector distance formed by digits

\begin{alignat*}{3}
  d_0 &&                         =  &&                distance(x_0, x_1) & \\
   d_1&&                      = &&              distance (x_0, x_2) &\\
\vdots &&      \vdots&&     \vdots & \\
 d_ i&&                      = &&              distance (x_3, x_7) &\\
\vdots &&      \vdots&&     \vdots & \\
   d_k&&                      = &&              distance (x_8, x_9) &\\
\end{alignat*}

In the second stage the algorithm is devided into two branches.
The feature set with labels from the model digit is fed directly to the MLP network and the weights are initialized.
The trainning set is devided into batches and using nearest neighbor algorithm, centroid and covariance matrix is calculated by 

\begin{equation}
S = \frac{1}{n-1} \sum_{i=1}^{n} \big( X_i -\bar{X}\big) \big( X_i -\bar{X}\big)' ,
\label{covar}
\end{equation}

from which the standard deviation is derived i.e. the square roots of the eigenvalues of the covariance matrix along the principal component. 
The centroid is simply the mean of the vectors.
if the distribution is spread, far more than the previously calculated standard deviation,  the weights of neural network is updated and varified against the testing set; consequently the accuracy over the test set  is calculated.

\begin{figure}[hhh]
    \centering
    \includegraphics[width=7.8cm]{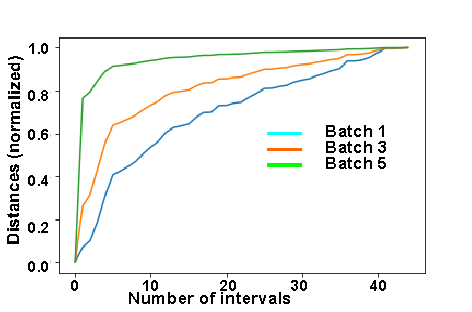}
    \caption{ Evolution of distribution of distances between intervals as a function of number of batches fed to the network.
      }
    \label{fig_5}
\end{figure}

Fig. \ref{fig_5} shows the evolution of the separation between feature vectors as the weights are updated with trainning batches.
The separation between feature set grows more distant during the trainning phase and increasing the accuracy.

 \begin{figure}[hhh]
    \centering
    \includegraphics[width=7.8cm]{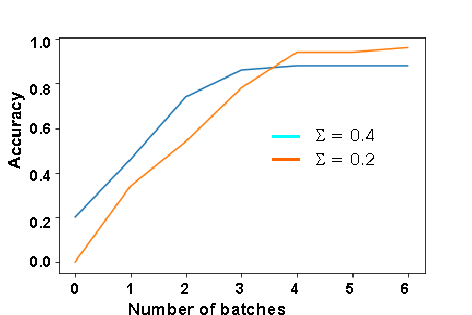}
    \caption{Accuracy as a function of batches for two different values of sigma deviation.
      }
    \label{fig_6}
\end{figure}

MNIST provides large data set for testing hence one can calculate accuracy with a low uncertainty.
In this case it was observed that the accuracy on testing data set depends on weight updating. 
If we allow for a larger Gaussian spread, the accuracy rises quickly over number of batches but does not reach its heighest value.
Whereas for smaller values about $98\%$ accuracy was acheived at the cost of larger number of batches.

The calculations were performed using computer cluster with 8 nodes consisting 2 CPUs each.
The module called \emph{mpi4py} was used for parallel processing \cite{mpi4py}.
The calculation of \emph{Centroid} and standard deviation was done using \emph{Scikit-learn} along with \emph{Numpy} library {Scikit-learn}.
The neural network was designed with the module known as \emph{Multi layer Perceptron } under \emph{Scikit-learn}.

\section {Conclusions}
\label {Conclusions}

A model fusing two different methods is presented here.
The performance of Gabor filter in image segmentation is also investigated and proved useful, though substantial amount of efforts were  required to fine-tune filter parameters for the output consistancy.
With this method we have achieved about $98\%$ accuracy. 
A sub-module is proposed to manipulate the images in order to align properly a distrorted image.
Early results have have shown increased efficiency of $99.2\%$.
Although this code is tested  only with the MNIST dataset,  the ongoing efforts  will use  the other datasets like Omniglot character sets and other abstract objects.

\end{document}